\documentclass[11pt]{amsart}
\usepackage{geometry}                
\usepackage{amsmath}
\usepackage{amssymb}
\usepackage{graphicx}
\usepackage{tikz-cd}
\usepackage{hyperref}
\usepackage[all]{xy}
\usepackage{tikz}
\usepackage{xspace}
\usepackage{bm}
\usepackage{amsmath}
\usepackage{amstext}
\usepackage{amsfonts}
\usepackage[mathscr]{euscript}
\usepackage{amscd}
\usepackage{latexsym}
\usepackage{amssymb}
\usepackage{enumerate}
\usepackage{xcolor}
\usepackage{mathtools}                
\usepackage{graphicx}
\usepackage{amssymb}
\usepackage{epstopdf}

\newcommand{\secat}{{\sf{secat}}}

\newcommand{\TC}{{\sf{TC}}}

\newcommand{\rr}{\mathbb{R}}

\newtheorem{theorem}{Theorem}[section]

\newtheorem{lemma}[theorem]{Lemma}

\newtheorem{definition}[theorem]{Definition}

\title{Sequential parametrized motion planning and its complexity, II}
\author{Michael Farber}
	\author{Amit Kumar Paul}
		\address{School of Mathematical Sciences\\
Queen Mary University of London\\ E1 4NS London\\UK.}
	
	\email{m.farber@qmul.ac.uk}
	
	\address{School of Mathematical Sciences\\
Queen Mary University of London\\ E1 4NS London\\UK.}
	
	\email{amit.paul@qmul.ac.uk/amitkrpaul23@gmail.com}
\date{\today}                                           

\begin{document}

	\thanks{Both authors were partially supported by an EPSRC grant}

	\begin{abstract} This is a continuation of our recent paper \cite{FarP} in which we developed the theory of {\it sequential parametrized} motion planning. 
A sequential parametrized motion planning algorithm produced a motion of the system which is required to visit a prescribed sequence of states, in a certain order, at specified moments of time. 
In \cite{FarP}  we analysed the sequential parametrized topological complexity of the Fadell - Neuwirth fibration which in relevant to the problem of moving multiple robots avoiding collisions with other robots and with obstacles in the Euclidean space. In  \cite{FarP} we found the sequential parametrised topological complexity of the Fadell - Neuwirth bundle for the case of the Euclidean space $\rr^d$ of odd dimension as well as the case $d=2$. In the present paper we give the complete answer for an arbitrary $d\ge 2$ even. Moreover, we present an explicit motion planning algorithm for controlling multiple robots in $\rr^d$ having the minimal possible topological complexity; this algorithm is applicable to any number $n$ of robots and any number $m\ge 2$ of obstacles. \end{abstract}
\maketitle
\section{Introduction}


 The topological approach to the motion planning problem of robotics \cite{Far03} centres around 
the notion of topological complexity $\TC(X)$, which has several different interpretations, see Theorem 14 in \cite{Far06}.  In particular, $\TC(X)$ is the minimal degree of instability of motion planning algorithms for systems having $X$ as their configuration space. 
A new \emph{\lq\lq parametrized\rq\rq} approach to the motion planning problem was developed recently in \cite{CohFW21}, \cite{CohFW}. Para\-metrized algorithms are universal and flexible, they are able to function 
in a variety of situations involving variable external conditions which are viewed as parameter and are part of the input of the algorithm. 

Generalizing this idea, the authors in \cite{FarP} 
developed the theory of \emph{sequential parametrized motion planning} in the spirit of Rudyak \cite{Rud10}; it involves an integer parameter $r\ge 2$ with the case $r=2$ reducing to the model of \cite{CohFW21}. 
In the sequential approach the robot requires 
to visit a given sequence of $r\ge 2$ states in certain order at prescribed moments of time. 

To make this paper readable we add a brief definition referring the reader to \cite{FarW} and \cite{FarP} for motivation and further detail. 
For a Hurewicz fibration $p : E \to B$ with fibre $X$ and an integer $r\ge 2$ we denote 
$$E^r_B= \{(e_1, \cdots, e_r)\in E^r; \, p(e_1)=\cdots = p(e_r)\}$$ 
and $E^I_B\subset E^I$ denotes 
the space of all paths $\alpha: I\to E$ such that $p\circ \alpha: I\to B$ is constant. Fix $r$ points 
$$0\le t_1<t_2<\dots <t_r\le 1$$ in $I$  (\lq\lq the time schedule\rq\rq)
and consider the evaluation map 
\begin{eqnarray}\label{Pir}
\Pi_r : E^I_B \to E^r_B, \quad \Pi_r(\alpha) = (\alpha(t_1), \alpha(t_2), \dots,  \alpha(t_r)).\end{eqnarray} 
$\Pi_r$ is a Hurewicz fibration, see \cite[Appendix]{CohFW}, the  fibre of $\Pi_r$ is $(\Omega X)^{r-1}$. 
A section $s: E^r_B \to E^I$ of the fibration $\Pi_r$ can be interpreted as {\it a parametrized sequential motion planning algorithm}, i.e. 
a function which assigns to every sequence of points $(e_1, e_2, \dots, e_r)\in E^r_B$ a continuous path $\alpha: I\to E$ (\lq\lq the motion of the system\rq\rq) satisfying $\alpha(t_i)=e_i$ for every $i=1, 2, \dots, r$ and such that the path 
$p\circ \alpha: I \to B$ is constant. The latter condition means that the system moves under constant external conditions 
(such as positions of the obstacles). 
Typically $\Pi_r$ does not admit continuous sections and hence the 
motion planning algorithms are necessarily discontinuous. 
The following definition gives a measure of complexity of sequential parametrized motion planning algorithms.

\begin{definition}\label{def:main}
The {\it $r$-th sequential parametrized topological complexity} of the fibration $p : E \to B$, denoted $\TC_r[p : E \to B]$, is defined as the sectional category of the fibration $\Pi_r$, i.e. 
\begin{eqnarray}\label{tcsec}
\TC_r[p : E \to B]:=\secat(\Pi_r).
\end{eqnarray}
\end{definition}	

In more detail, $\TC_r[p : E \to B]$ is the minimal integer $k$ such that there is a open cover 
$\{U_0, U_1, \dots, U_k\}$ of $E^r_B$ with the property that each open set $U_i$ 
admits a continuous section $s_i : U_i \to E^I_B$ of $\Pi_r$.
The following Lemma allows using arbitrary partitions instead of open covers: 

 \begin{lemma}[see Proposition 3.6 in \cite{ FarP}]\label{lemma para tc}
 Let $p: E \to B$ be a locally trivial fibration where $E$ and $B$ are metrisable separable ANRs. Then the $r$-th sequential parametrized topological complexity $\TC_r[p: E \to B]$ equals the smallest integer $k\ge 0$ such that the space $E_B^r$ admits a partition $$E_B^r=F_0 \sqcup F_1 \sqcup ... \sqcup F_k, \quad F_i\cap F_j= \emptyset \text{ \ for } i\neq j,$$
and on each set $F_i$ there exists a continuous section $s_i : F_i \to E_B^I$ of the fibration $\Pi_r$.
 \end{lemma}

The aim of this paper is to give an explicit sequential parametrized motion planning algorithm for 
the problem of moving $n\ge 1$ robots in the presence of $m\ge 2$ obstacles in $\rr^d$ where $d\ge2 $ is even. This problem can be expressed through the properties of the 
Fadell-Neuwirth bundle 
\begin{eqnarray}\label{fn}
p : F(\rr^d, m+n) \to F(\rr^d, m)
\end{eqnarray}
and its topological complexity $\TC_r[p : F(\rr^d, m+n) \to F(\rr^d, m)]$ as defined above, see (\ref{tcsec}). The notation  $F(\rr^d, m+n)$ stands for the configuration space of $m+n$ pairwise distinct points of $\rr^d$ labeled by the integers $1, 2, \dots, m+n$. The projection (\ref{fn}) associates with a configuration of $m+n$ points the first $m$ points. 

The main result of this paper is the following:

\begin{theorem}\label{thmain}
For any even $d\geq 2$ and for any integers $n\geq 1$, $m\geq 2$, $r\geq 2$, the sequential parametrized topological complexity of the Fadell-Neuwirth bundle (\ref{fn}) is given by 
\begin{eqnarray}\label{tcfn}
\TC_r[p : F(\rr^d, m+n) \to F(\rr^d, m)]=rn + m - 2.
\end{eqnarray}
\end{theorem}

In the case $d=2$ this statement was proven in \cite{FarP}, Theorem 9.2. 
Besides, Proposition 9.1 from 
\cite{FarP} gives in (\ref{tcfn}) the inequality $\ge$, which is valid for any $d\ge 2$ even. Hence to prove Theorem \ref{thmain} it is enough to establish for $\TC_r[p : F(\rr^d, m+n) \to F(\rr^d, m)]$
the upper bound $rn+m-2$. This will follow once we present an explicit sequential parametrized motion planning algorithm of complexity $\le rn + m - 2$. This task will be accomplished in \S \ref{sec:three} of this paper. 

In the special case of $r=2$ 
a motion planning algorithm for the Fadell - Neuwirth bundle  
for dimension $d$ odd  was given in \cite{FarW}; it has higher by one topological complexity. 
We refer also to \cite{GonZ} where the case $r=2$ and $m=2$  (two obstacles) was considered.

\section{The obstacle avoiding manoeuvre}\label{man} 
In this section we describe an obstacle avoiding manoeuvre which will be used below as a sub-algorithm in the general algorithm. We shall consider the situation of a single robot moving avoiding collisions with $m\ge 2$ obstacles in $\rr^d$ where the dimension $d\ge 2$ is assumed to be even. 

Let $m\ge 2$ be an integer and let $R$ be an equivalence relation on the set $[m]=\{1, 2, \dots, m\}$. For $i, j\in [m]$ we shall write $i\sim_R j$ to indicate that $i$ and $j$ are equivalent with respect to $R$. We shall denote by $m(R)\le m$ the number of equivalence classes of $R$. 

Consider the configurations 
\begin{eqnarray}\label{conf0}
C= (o_1, o_2, \dots, o_m, z, z'),
\end{eqnarray} 
where $o_i\in \rr^d$ denote positions of the obstacles and the points $z, z'\in \rr^d$ denote the current and the desired positions of the robot, correspondingly. Denote by 
$$b=(o_1, \dots, o_m)$$ the configuration of the obstacles. Assuming that $o_1\not=o_2$ we can consider 
the unit vector $$e_b=||o_2-o_1||^{-1}\cdot (o_2-o_1)$$ 
and the line $L_b\subset \rr^d$ through the origin which is parallel to $e_b$. 
We shall denote by 
$q_b: \rr^d\to L_b$ the orthogonal projection onto the line $L_b$, i.e. $q_b(x)=\langle x, e_b\rangle\cdot e_b$.

Given an equivalence relation $R$ on $[m]$ we define 
the space $$\Omega_{m, R}$$ as the set of all configurations (\ref{conf0}) satisfying the following conditions:

(a) $o_i\not= o_j$ and $z\not= o_i\not=z'$ for all $i, j\in [m], \, i\not =j$; 

(b) $q_b(o_i)=q_b(o_j)$ if and only if $i\sim_R j$ where $i, j\in [m]$; 

(c) $q_b(z)\not=q_b(o_i)\not=q_b(z')$ for $i\in [m]$. 

For a configuration $C$ as in (\ref{conf0}) we shall denote by
$\epsilon(C)>0$ the minimum of the numbers 
\begin{eqnarray*}
&|o_i-o_j|, \quad i\not=j,\quad i, j\in [m],\\ 
&|q_b(z)- q_b(o_i)|,\quad |q_b(z')- q_b(o_i)|,\quad i\in [m],\\
&|q_b(o_i)-q_b(o_j)|, \quad i\not \sim_R j, \quad i, j \in [m]. 
\end{eqnarray*}
Note that $\epsilon(C)$ is a continuous function of $C\in \Omega_{m,R}$.

Our goal is to describe an explicit function associating with each configuration $C\in \Omega_{m, R}$ a continuous path
$\gamma_C: I=[0,1]\to \rr^d$ such that
\begin{eqnarray}\label{goal}
\gamma_C(0)=z, \quad \gamma_C(1)=z', \quad  \gamma_C(t)\not=o_i\quad\mbox{ for any}\quad  t\in I, \quad 
i\in [m].\end{eqnarray} 
Moreover, we require the function $\gamma_C(t)$ to be continuous function of two variables $(C, t)\in \Omega_{m,R}\times I$.

Clearly $\Omega_{m,R}$ is the disjoint union 
$$\Omega_{m,R}=\sqcup_{j=0}^{m(R)} \Omega^j_{m,R}$$ where $\Omega^j_{m,R}\subset \Omega_{m,R}$ denotes the set of configurations (\ref{conf0}) such that exactly $j$ distinct projection points 
$q_b(o_i)$ lie between the projections $q_b(z)$ and $q_b(z')$ onto $L_b$. 

Each of the subsets $\Omega^j_{m,R}$ is open and closed in $\Omega_{m,R}$ and hence in it enough to define $\gamma_C(t)$ for $C\in \Omega^j_{m,R}$, where $j=0, 1, \dots, m(R)$ and $R$ are fixed.

For $C\in \Omega^0_{m,R}$ we can simply define the $\gamma_C(t) = (1-t)z+tz'$, where $t\in [0,1]$.

Next we consider the case when there are $j\ge 1$ projections of the obstacles $q_b(o_i)$ lying between the projections $q_b(z)$ and $q_b(z')$.  For simplicity we shall assume that $q_b(z)< q_b(z')$ and denote
$$q_b(z)< q_b(o_{i_1})< \dots< q_b(o_{i_j})< q_b(z').$$
Here each $i_s\in [m]$ is the smallest index in its $R$-equivalence class.

We introduce the points
\begin{eqnarray}
o_{i_s}^\pm=o_{i_s}\pm \frac{1}{4}\epsilon(C) \cdot e_b\, \in \rr^d, \quad s=1, \dots, j. 
\end{eqnarray}
These points are small perturbations of the positions of the obstacles in the direction parallel to the line $L_b$ . 

Recall that the dimension $d$ of the space is even and hence there exists a continuous function associating with any unit vector $e\in \rr^d$ a unit vector $e^\perp\in \rr^d$ which is perpendicular to $e$. Such function $e\mapsto e^\perp$ is a tangent vector field of the sphere $S^{d-1}$. 

\begin{figure}[h]
\begin{center}
\includegraphics[scale=0.4]{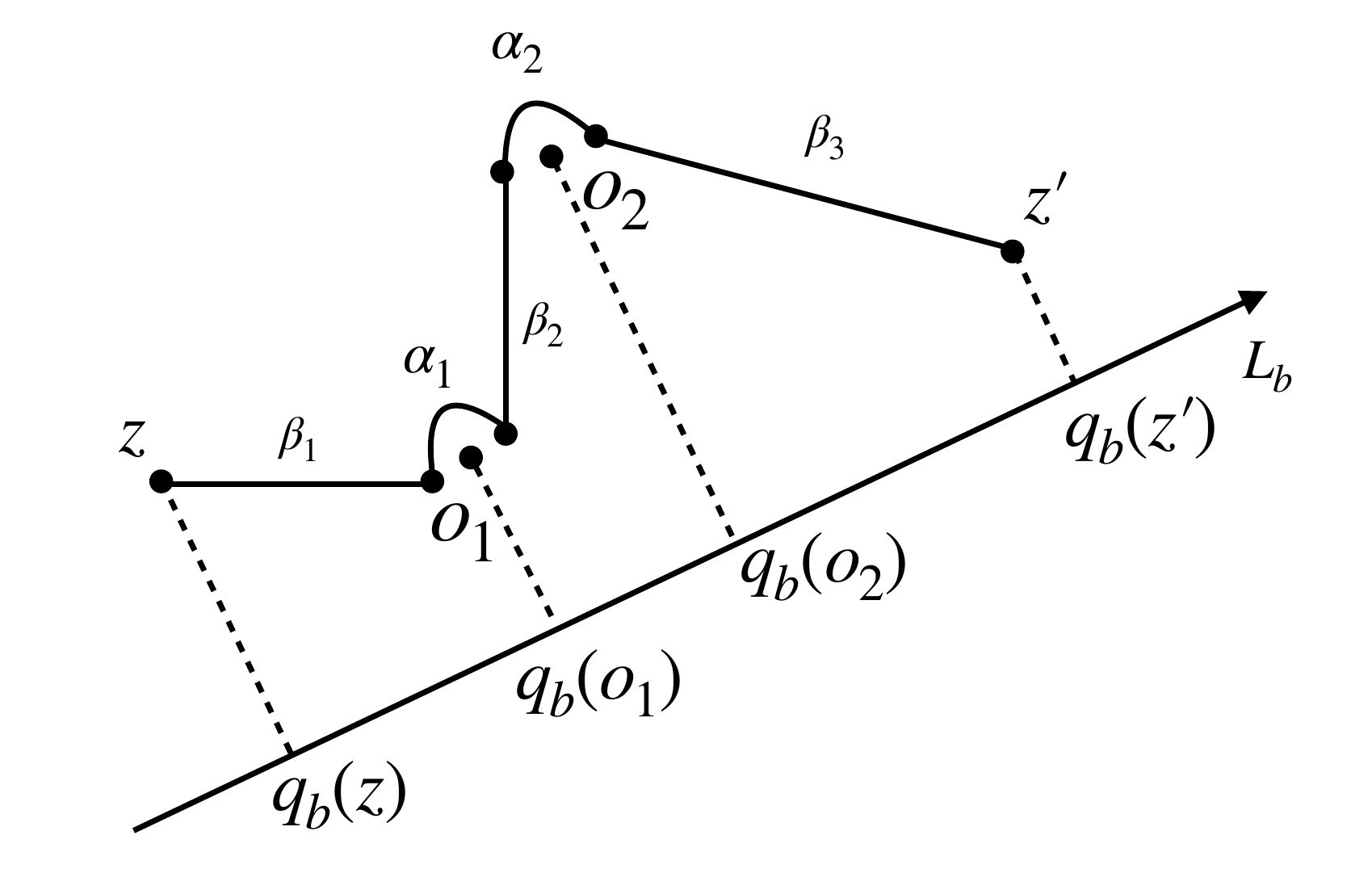}
\caption{Motion of the robot from $z$ to $z'$ avoiding collisions with obstacles.}
\label{motion}
\end{center}
\end{figure}

For $s=1, 2, \dots, j$
the path
\begin{eqnarray}
\alpha_s(t) = o_{i_s} - \frac{\epsilon(C)}{4}\cdot [\cos(\pi t)\cdot e_b + \sin(\pi t)\cdot e_b^\perp], \quad t\in [0,1],
\end{eqnarray}
(semi-circle) connects the point $o^-_{i_s}$ to $o^+_{i_s}$ avoiding the obstacles. The straight line path 
$$\beta_s(t) = (1-t)\cdot o^+_{i_{s-1}} + t\cdot o^-_{i_s}, \quad s=2,\dots, j$$
connects the point $o^+_{i_{s-1}}$ to $o^-_{i_s}$ avoiding the obstacles. We can similarly define the straight line path 
$\beta_1(t)= (1-t)\cdot z+ t\cdot o^-_{i_1}$ and $\beta_{j+1}= (1-t)\cdot o^+_{i_j}+t\cdot z'$. 
The concatenation 
\begin{eqnarray}
\gamma_C = \beta_1\ast \alpha_1\ast \beta_2\ast\alpha_2\ast\dots\ast \alpha_j \ast\beta_{j+1}
\end{eqnarray}
(consisting of the straight line segments and semi-circles) is the desired path. Thus we obtain a continuous function $ \Omega_{m, R}\times I\to \rr^d$, $(C,t)\mapsto \gamma_{C}(t)$ satisfying (\ref{goal}).

\section{Collision-free motion planning algorithm for many robots and obstacles in $\rr^d$; the even-dimensional case}\label{sec:three}

In this section we describe an explicit parametrized sequential motion planning algorithm for collision free motion of an arbitrary number $n$ of robots in the presence of an arbitrary number of $m$ moving obstacles in $\rr^d$, where the dimension $d\ge 2$ is assumed to be even. This algorithm is optimal for $d\ge 2$ even in the sense that its topological complexity is minimal possible and coincides with the cohomological lower bound established in \cite{FarP}. 

For the special case $r=2$ a motion planning algorithm of this kind was described in \S 6 of \cite{FarW} which is applicable for any dimension $d\ge 2$, but its topological complexity is minimal only for $d$ odd. 
The algorithm described in this section is only applicable for even $d\ge 2$, it has a smaller number of local rules and it is {\it sequential}, i.e. it is applicable for any $r\ge 2$.  

We denote by $p: E\to B$ the Fadell - Neuwirth bundle 
\begin{eqnarray}\label{fadell}
p: F(\rr^d, m+n)\to F(\rr^d, m)\end{eqnarray} 
where $F(\rr^d, m+n)$ is the configuration space of $m+n$ pairwise distinct points of $\rr^d$.
A point of $E=F(\rr^d, m+n)$ will be denoted by the symbol 
$(o_1, o_2, ..., o_m, z_1, z_2, ..., z_n)$, where $o_i\in \rr^d$ are positions of $m$ obstacles and $z_j$'s are positions of $n$ robots. The projection (\ref{fadell}) acts as follows
$$p(o_1, o_2, ..., o_m, z_1, z_2, ..., z_n)= (o_1, o_2, ..., o_m) \in B = F(\rr^d, m).$$
The map (\ref{fadell}) is a locally trivial bundle, see \cite{FadN62}.

Let us consider the fibration 
\begin{eqnarray}\label{fib}
\Pi_r: E^I_B \to E^r_B\end{eqnarray} 
built out of the Fadell - Neuwirth bundle (\ref{fadell}). 
A point of the space $E_B^r$ can be viewed as a configuration
\begin{eqnarray}\label{conf}
C=(o_1, o_2, ..., o_m, z^1_1, z^1_2, ..., z^1_n, z^2_1, z^2_2, ..., z^2_n,..., z^r_1, z^r_2, ..., z^r_n),\end{eqnarray} 
where $o_i, z_j^\ell\in \rr^d$ are such that: 
\begin{enumerate}

\item $o_i \neq o_{i'}$ for $i\neq i'$,

\item $o_i \neq z^\ell_j$ for $1\leq i \leq m$,  $1\leq j \leq n$ and $1\leq \ell \leq r$,

\item $z^\ell_j \neq z^\ell_{j'}$ for $j \neq j'$ and $1\leq \ell \leq r$.
\end{enumerate}

\subsection{Partition of the space $E^r_B$} For a configuration $C\in E^r_B$ as in (\ref{conf}) 
let $b$ denote the configuration $b=(o_1, o_2, ..., o_m) \in B= F(\rr^d, m)$, and let $L_b\subset \rr^d$ denote the line passing through the origin and the vector $o_2-o_1\in \rr^d$. We shall orient the line $L_b$ such that the vector $o_2-o_1$ points in the positive direction. Let $e_b\in L_b$ denote the unit vector in the direction of the orientation. 

It is well known that for even $d\ge 2$ the unit sphere $S^{d-1}\subset \rr^d$ admits a continuous non-vanishing tangent vector field. Fixing such a vector field gives a continuous function $e\mapsto e^\perp$ which assigns to every unit vector 
$e\in \rr^d$ a unit vector $e^\perp\in \rr^d$ perpendicular to $e$. The vector $e^\perp$ will help us to specify the obstacle 
avoiding motions. 

For $x\in \rr^d$, we denote by $q_b(x)$ the orthogonal projection of $x$ onto $L_b$, i.e. $q_b(x) =\langle x, e_b\rangle\cdot e_b$ where $\langle \, , \, \rangle$ denotes the scalar product. 

For a configuration (\ref{conf}) we denote by
$$q_b(C)=\{q_b(o_i), q_b(z_j^\ell) \ | \ 1\leq i \leq m, 1\leq j \leq n, 1\leq \ell \leq r \}$$
the set of all projections of the obstacles and robots onto the line $L_b$. 
Note that the map $q_b$ depends on the positions of the obstacles $o_1$ and $o_2$ of the configuration $C$. 
Clearly, $q_b(o_1)< q_b(o_2)$; thus, the cardinality of the set $q_b(C)$ satisfies the relation 
$$2\leq |q_b(C)| \leq rn+m.$$ 
For any $2\leq c \leq rn+m$ we define the subset $W_{c}\subset E^r_B$ as follows
\begin{eqnarray*}\label{two}
W_{c}=\{C\in E_B^r \ | \ q_b(C) \text{ has cardinality } c\}.
\end{eqnarray*}
Clearly, 
\begin{eqnarray}\label{one1}
E_B^r = W_2 \sqcup W_3 \sqcup ... \sqcup W_{rn+m}.\end{eqnarray}
Below we construct a continuous section of the fibration $\Pi_r: E^I_B \to E^r_B$ over each set $W_c$, where $c=2, 3, \dots, rn+m$. The closure of $W_c$ satisfies
\begin{eqnarray*}
\overline W_c \subset \bigcup_{c'\le c} W_{c'}. 
\end{eqnarray*}

\subsection{Partitioning the set $W_{c}$.} Next we partition $W_c$ as follows:
\begin{eqnarray}\label{one}
W_{c}=\bigsqcup_{s+t=c} G_{s, t},
\end{eqnarray}
where $G_{s, t}\subset W_c$ is defined as the set of all configurations (\ref{conf}) such that the set \newline
$\{q_b(o_1), q_b(o_2), ..., q_b(o_m)\}$ has cardinality $s$. Clearly, $s$ takes values $2, 3, \dots, m$. The index $t$ is defined by the relation $s+t=c$; the value of $t$ equals the number of distinct projection values $q_b(z_j^\ell)\in L_b$ which are not projections of the obstacles $q_b(o_i)$. 
The closure of $G_{s, t}$ satisfies 
\begin{eqnarray}\label{one2}
\overline G_{s,t} \subset \bigcup_{s'\le s,\, \, \, t'\le t} G_{s', t'}.
\end{eqnarray}
This implies that each set $G_{s,t}$ is open and closed as a subset of $W_c$, where $c=s+t$, and hence continuous sections over the sets $G_{s,t}$ defined below give jointly a continuous section over $W_c$. 

%
%
%

\subsection{Partitioning the sets $G_{s, t}$.} Consider the set of $m+rn$ formal symbols 
\begin{eqnarray}\label{set}
S= \{o_1, \dots, o_m, z_1^1, \dots, z^1_n, \dots, 
z_1^r, \dots, z^r_n\}.\end{eqnarray}
 A configuration (\ref{conf}) defines {\it a linear quasi-order} on the set $S$, where for $s, s'\in S$ we say that $s\le s'$ iff $q_b(s)\le q_b(s')$. 
A quasi-order allows for distinct elements $s, s'\in S$ to satisfy $s\le s'$ and $s'\le s$. In such a case we shall say that the elements $s, s'\in S$ are {\it equivalent with respect to this quasi-order}. 

Consider the set $\mathcal O^{m+rn}_{s, t}$ of all possible quasi-orders on the set $S$
having in total $s+t$ equivalence classes and such that the set $\{o_1, \dots, o_m\}$ has $s$ equivalence classes. 
For $\sigma \in \mathcal O^{m+rn}_{s, t}$, 
let $A^\sigma_{s, t}$ denote the set of all configurations $C\in G_{s, t}$ generating the quasi-order $\sigma$ on the set $S$ of symbols (\ref{set}). 
Clearly, one has 
\begin{eqnarray}\label{asigma}\label{one3}
G_{s, t} =\bigsqcup_{\sigma\in \mathcal O^{m+rn}_{s, t}} A^{\sigma}_{s, t}, \quad \mbox{where}\quad s\ge 2.\end{eqnarray} 
Each of the sets $A^{\sigma}_{s, t}$ is open and closed in $G_{s, t}$ hence a collection of continuous sections over $A^{\sigma}_{s, t}$ (with various quasi-orders $\sigma$)  define together a continuous section over 
$G_{s, t}$. 

\subsection{ Motion planning algorithm on $A^{\sigma}_{s, rn}$.}\label{sec24} In this subsection we describe a continuous section of the fibration (\ref{fib}) over $A^{\sigma}_{s, rn}\subset E^r_B$.
For a configuration (\ref{conf}) lying in $A^\sigma_{s,rn}$
all projection points $q_b(z_i^\ell)\in L_b$ are pairwise distinct, where $i=1, \dots, n$ and $\ell=1, \dots r$; these points are also distinct from the projections of the obstacles
$q_b(o_j)$ where $j=1, \dots, m$. In general, $s\le m$, and hence equality  $q_b(o_i)=q_b(o_j)$ for $i\not= j$ is not excluded; this happens iff the numbers $i$ and $j$ are equivalent with respect to the quasi-order $\sigma$. 
A section of (\ref{fib}) over $A^{\sigma}_{s, rn}$ is determined by 
 $n$ functions 
\begin{eqnarray}\label{curves}
\gamma_C^1, \, \gamma_C^2,\,  \dots, \, \gamma_C^n: \, I\to \rr^d
\end{eqnarray}
satisfying the following conditions: 
\begin{itemize}
\item[{(a)}] Each $\gamma_C^i(t)$ is continuous as a function of $(C, t)\in A^\sigma_{s,rn}\times I$ for $i=1, \dots, n$; 
\item[{(b)}] $\gamma^i_C(t) \not= \gamma^j_C(t)$, for $i\not=j$ and $t\in I$; 
\item[{(c)}] $\gamma^i_C(t) \not= o_j$ for $i=1, \dots, n$,  $j=1,\dots, m$ and $t\in I$; 
\item[{(d)}] $\gamma_C^i(t_\ell) = z^\ell_i$ where $\ell=1, \dots, r$ and $i=1, \dots, n$. 
\end{itemize}
In (d) the symbols $t_\ell\in I$ denote the fixed time moments $$0=t_1<t_2 < \dots< t_r=1$$ used in the definition of the evaluation map (\ref{fib}). The section $s$ determined by the system of curves (\ref{curves}) is given by 
$$s(C)=(o_1, \dots, o_m, \gamma_C^1, \dots, \gamma_C^n)\in E^I_B.$$

To construct the curves (\ref{curves}) we shall use the obstacle avoiding manoeuvre of \S \ref{man}.
Divide each time interval $[t_{\ell}, t_{\ell+1}]$, where $\ell=1, \dots, r-1$, into $n$ subintervals 
$$t_\ell=t_{\ell, 0}<t_{\ell, 1}< \dots< t_{\ell, n}=t_{\ell+1}$$
and add to properties (a) - (d) an additional requirement:
\begin{itemize}
\item[{(e)}] the curve 
$\gamma_C^i(t)$ is not constant only for $t\in [t_{\ell, i-1}, t_{\ell, i}]$ where $\ell=1, \dots, r-1$.  
\end{itemize}

Next we describe the system of curves (\ref{curves}) satisfying the conditions (a) - (e). 
On the first interval $[t_{1,0}, t_{1,1}]$ we apply the algorithm of \S \ref{man} to move the first robot from $z_1^1$ to $z_1^2$ viewing $o_1, \dots, o_m, z_2^1, z_3^1, \dots, z_n^1$ as obstacles. Note that the assumptions of \S \ref{man} are satisfied since $C\in A^\sigma_{s, rn}$.
This defines the curve $\gamma_C^1|_{[t_{1,0}, t_{1,1}]}$. 
For $i=2, \dots, n$ the curve
$\gamma_C^i|_{[t_{1,0}, t_{1,1}]}$ is the constant curve at $z_i^1$.

On the next interval $[t_{1,1}, t_{1,2}]$ we apply the algorithm of \S \ref{man} to move the second robot from
$z_2^1$ to $z_2^2$ viewing $o_1, \dots, o_m, z_1^2, z_3^1, \dots, z_n^1$ as obstacles. This defines the curve $\gamma_C^2|_{[t_{1,1}, t_{1,2}]}$.
For $i=1$ the curve
$\gamma_C^i|_{[t_{1,1}, t_{1,2}]}$ is the constant curve at $z_1^2$ while for 
$i=3, \dots, n$ the curve
$\gamma_C^i|_{[t_{1,1}, t_{1,2}]}$ is the constant curve at $z_i^1$.

Continuing in this fashon we construct the curves $\gamma_C^i|_{[t_{\ell, j-1}, t_{\ell, j}]}$ inductively using the algorithm of \S \ref{man} as follows. On the time interval $[t_{\ell, j-1}, t_{\ell, j}]$ the robot $j$ moves from the position 
$z_j^\ell$ to $z_j^{\ell+1}$ and this motion is specified by the algorithm of \S \ref{man} in which we consider the points $o_1, \dots, o_m, z_1^{\ell+1}, \dots, z_{j-1}^{\ell+1}, z_{j+1}^\ell, \dots, z_n^\ell$ as obstacles. This defines the curve $\gamma_C^j|_{[t_{\ell, j-1}, t_{\ell, j}]}$. Besides, for $i<j$ the curves 
$\gamma_C^i|_{[t_{\ell, j-1}, t_{\ell, j}]}$ is constant at $z_i^{\ell+1}$ while for $i>j$ the curve $\gamma_C^i|_{[t_{\ell, j-1}, t_{\ell, j}]}$ is constant at $z_i^\ell$.

\subsection{Desingularization: motion planning algorithms on the sets $A^{\sigma}_{s, t}$ with $t<rn$.} 
\begin{figure}[h]
\begin{center}
\includegraphics[scale=0.35]{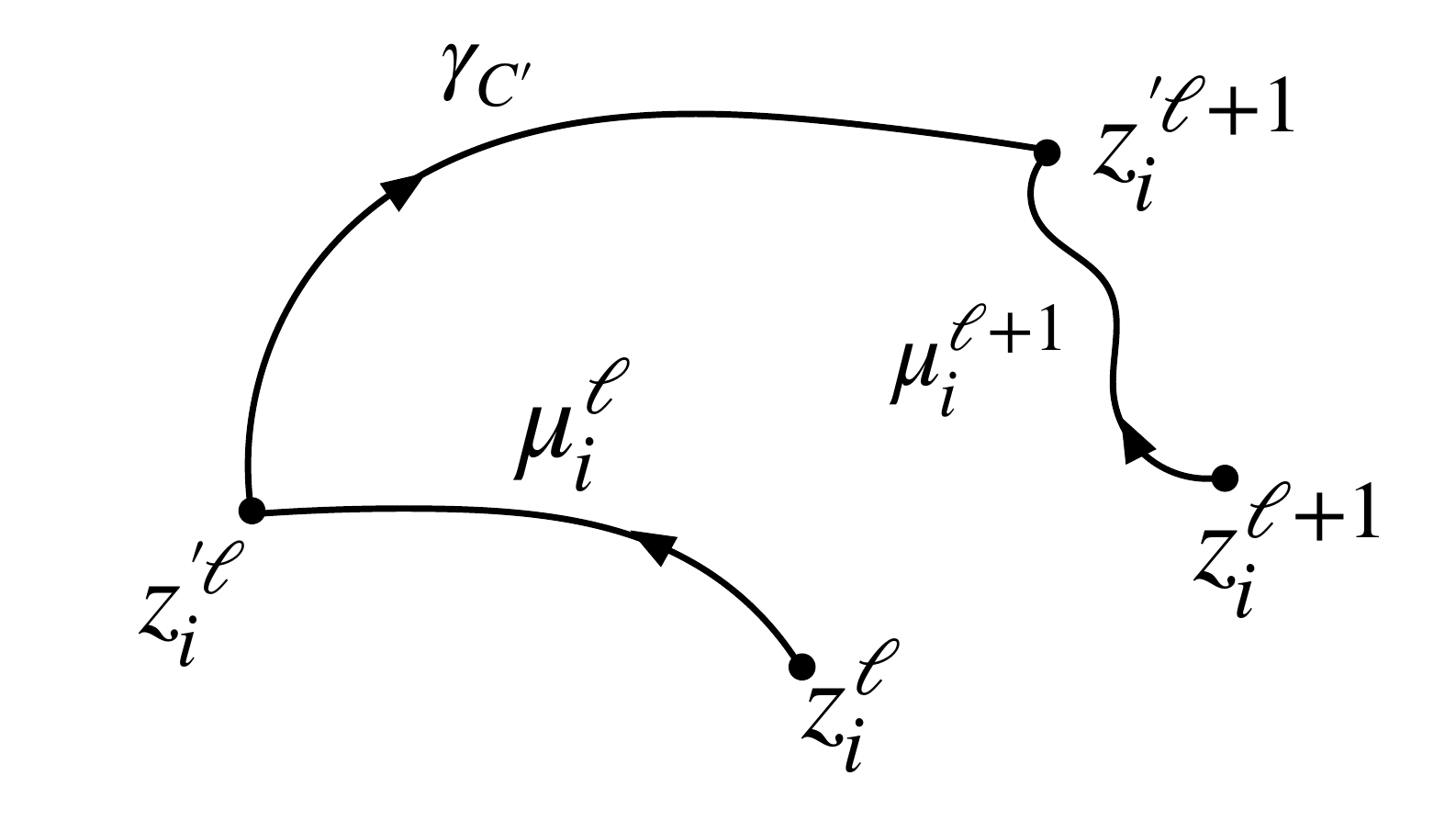}
\caption{Motion of the robot from $z_i^\ell$ to $z_i^{\ell+1}$.}
\label{general}
\end{center}
\end{figure}

For a configuration $C\in A^\sigma_{s,t}$ define the number $\delta(C)>0$ as the minimum of the numbers 
$$
\begin{array}{lll}
|q_b(z^\ell_i)-q_b(z^{\ell'}_k)|,& \mbox{where} & q_b(z^\ell_i)\not= q_b(z^{\ell'}_k),\\ \\
|q_b(z^\ell_i) -q_b(o_j)|, & \mbox{where} & q_b(z^\ell_i) \not= q_b(o_j),  
\end{array}
$$
where $i, k \in \{ 1, \dots, n\}$, $\ell, \ell' \in \{1, \dots, r\}$ and $j\in \{1, \dots, m\}$. Note that $\delta(C)$ is a continuous function of $C\in A^\sigma_{s,t}$. 
Next we define a homotopy $H: A^{\sigma}_{s, t} \times I \to E^r_B$ as follows.
For $C\in A^\sigma_{s,t}$ we set 
$$
H(C,t)= (o_1, \dots, o_m, \mu_1^1(t), \dots, \mu_n^1(t), \dots, \mu_1^r(t), \dots, \mu^r_n(t))\in E^r_B
$$
where 
$$\mu_i^\ell(t)=z_i^\ell + \frac{((\ell-1)n+i)\cdot t \cdot \delta(C)}{rn}\cdot e_b.$$
Note that for $t>0$ the configuration $H(C,t)$ lies in $A_{s, rn}^\tau$. Here $\tau$ is the following quasi-ordering on the set (\ref{set}): 
\begin{itemize}
\item [{(a)}] if for two symbols $s, s'$ of (\ref{set}) one has $s\le_\sigma s'$ then 
$s\le_{\tau} s'$; 
\item [{(b)}] if two distinct symbols $z_i^\ell$ and $z^{\ell'}_{i'}$ of (\ref{set}) are equivalent with respect to $\sigma$ then $z_i^\ell <_{\tau} z^{\ell'}_{i'}$ assuming that $(\ell-1)n+i<(\ell'-1)n+i'$. 
\item [{(c)}] if two symbols $o_j$ and $z_i^\ell$ of (\ref{set}) are equivalent with respect to the quasi-order 
$\sigma$ then 
$o_j <_{\tau} z_i^\ell$. 
\end{itemize}
Denoting $z_i^\ell(1)={z'}_i^\ell$ we have the configuration
$$
C'=(o_1, \dots, o_m, {z'}_1^1, \dots, {z'}_n^1, \dots, {z'}_1^r, \dots {z'}_n^r)\in A^{\tau}_{s, rn}
$$
to which we may apply the algorithm of section \ref{sec24}. 
The path $\mu_i^\ell(t)$
 connects the points $z_i^\ell$ and ${z'}_i^\ell$.
 We obtain a path connecting $z_i^\ell$ to $z_i^{\ell +1}$ 
 as a concatenation of $\mu_i^\ell$ with the path $\gamma_{C'}$ from ${z'}_i^\ell$ to ${z'}_i^{\ell+1}$ given the algorithm of \S \ref{sec24} followed by the inverse of the path $\mu_i^{\ell+1}$, see Figure \ref{general}. 

	\section{Proof of Theorem \ref{thmain}}
	As we mentioned above, the inequality $\TC_r[p : F(\rr^d, m+n) \to F(\rr^d, m)]\ge rn+m-2$ is given by 
Proposition 9.1 from 
\cite{FarP}. To obtain the opposite inequality we shall use Lemma \ref{lemma para tc} applied to the partition
(\ref{one1}). The construction of \S \ref{sec:three} defines a continuous section of the bundle (\ref{fib}) over each of the sets $W_c$, where $c=2, 3, \dots, rn+m$. Indeed, we described in \S \ref{sec:three} a continuous section over each of the sets $A^\sigma_{s, t}$. The sets $A^\sigma_{s, t}$ are open and closed in $G_{s, t}$, see (\ref{one3}), and the sets $G_{s, t}$ are open and closed in $W_c$, where $c=s+t$, see (\ref{one}). 
Thus, the continuous sections over the sets $A^\sigma_{s, t}$ with $s+t=c$ jointly define a continuous section over the set $W_c$. 
This completes the proof. 

%


\begin{thebibliography}{12}
		
		
		
				
		
		
		
		\bibitem{CohFW21} D.C.~ Cohen, M.~ Farber and S.~ Weinberger, ``Topology of parametrized motion planning algorithms," \textit{J. Appl. Algebra Geometry} \textbf{5}, (2021), no. 2, pp.~229 - 249.
		
\bibitem{CohFW} D.C.~ Cohen, M.~ Farber and S.~ Weinberger, ``Parametrized topological complexity of collision-free motion planning in the plane," 		
		Annals of Mathematics and Artificial Intelligence,  90(2022), 999-1015.
		
		\bibitem{FadN62} E.~ Fadell and L.~ Neuwirth,  \textit{Configuration spaces.} Math. Scand. \textbf{10} (1962), pp.~111 - 118.
		
		
		
		\bibitem{Far03} M.~ Farber, ``Topological complexity of motion planning," \textit{Discrete Comput. Geom.} \textbf{29} (2003), no. 2, pp.~211 - 221.
		\bibitem{Far06} M. Farber, \emph{Topology of robot motion planning}. Morse theoretic methods in nonlinear analysis and in symplectic topology, 185–230, NATO Sci. Ser. II Math. Phys. Chem., 217, Springer, Dordrecht, 2006. 
		
		\bibitem{FarP} M. Farber and A.K. Paul, \textit{Sequential parametrized motion planning and its complexity}. Topology Appl. 321 (2022), Paper No. 108256, 23 pp.


	

	
\bibitem{FarW} M.~ Farber and S.~ Weinberger, ``Parametrized motion planning and topological complexity," \textit{	arXiv:2202.05801}, (2022). To appear in Proceeding of the Workshop on Algorithmic Foundation of Robotics WAFR22. 



%
		
		
		
	
		  
		
		
		
		
		
		
		\bibitem{GonZ} C. A.  Zapata and J. González, \textit{Parametrised collision-free optimal motion planning algorithms in Euclidean spaces}, arXiv:2103.14074 


 
		
		\bibitem{Rud10} Y. B. ~Rudyak, `` On higher analogs of topological complexity," \textit{Topology Appl.} \textbf{157} (2010), no. 5, pp.~916 - 920.


		
		
		 
		
	\end{thebibliography}
\end{document}